\documentclass[11pt]{article}
\usepackage[margin=1in]{geometry}
\usepackage{graphicx}  % Image handling
\usepackage{amsmath, amssymb}  % Math support
\usepackage[colorlinks=true, allcolors=blue, breaklinks=true]{hyperref}
\usepackage{microtype}  % Better spacing
\usepackage{subfig}  % Instead of deprecated subfigure
\usepackage{overpic}  % Overlay figures
\usepackage{booktabs}  % For professional tables
\usepackage{multirow}  % Multi-row table cells
\usepackage{adjustbox}  % Adjust box sizes
\usepackage{authblk}  % Multi-author formatting
\usepackage[numbers]{natbib}
\bibpunct{[}{]}{,}{n}{}{;}
\usepackage{hyperref}

\title{Low-Resource Video Super-Resolution using Memory, Wavelets, and Deformable Convolutions}

\author[1]{Kavitha Viswanathan}
\author[1]{Amit Sethi}
\author[1]{Shashwat Pathak}
\author[1]{Piyush Bharambe}
\author[1]{Harsh Choudhary}
\affil[1]{Department of Electrical Engineering, IIT Bombay, India}

\begin{document}
\maketitle
\footnote{Code is available at: \url{https://github.com/kavi1388/RCDM}}
\begin{abstract}
The tradeoff between reconstruction quality and compute required for video super-resolution (VSR) remains a formidable challenge in its adoption for deployment on resource-constrained edge devices. While transformer-based VSR models have set new benchmarks for reconstruction quality in recent years, these require substantial computational resources. On the other hand, lightweight models that have been introduced even recently struggle to deliver state-of-the-art reconstruction. We propose a novel lightweight and parameter-efficient neural architecture for VSR that achieves state-of-the-art reconstruction accuracy with just 2.3 million parameters. Our model enhances information utilization based on several architectural attributes. Firstly, it uses 2D wavelet decompositions  strategically interlayered with learnable convolutional layers to utilize the inductive prior of spatial sparsity of edges in visual data. Secondly, it uses a single memory tensor to capture inter-frame temporal information while avoiding the computational cost of previous memory-based schemes. Thirdly, it uses residual deformable convolutions for implicit inter-frame object alignment that improve upon deformable convolutions by enhancing spatial information in inter-frame feature differences. Architectural insights from our model can pave the way for real-time VSR on the edge, such as display devices for streaming data.
\end{abstract}
%Transformer-based video super-resolution (VSR) models have set new benchmarks for reconstruction quality in recent years, but their substantial computational demands make most of them unsuitable for deployment on resource-constrained devices. Achieving a balance between model complexity and output quality remains a formidable challenge in VSR. Although lightweight models have been introduced to address this issue, they often struggle to deliver state-of-the-art reconstruction. We propose a novel lightweight, parameter-efficient deep residual deformable convolution network for VSR. Unlike prior methods, our model enhances feature utilization through residual connections and employs deformable convolution for precise frame alignment, addressing motion dynamics effectively. Furthermore, we introduce a single memory tensor to capture information accrued from the past frames and improve motion estimation across frames along with a wavelet-based enhancement to strengthen frequency content in super-resolved frames. This design enables an efficient balance between computational cost and reconstruction quality. With just 2.3 million parameters, our model achieves state-of-the-art SSIM of 0.9175 on the REDS4 dataset, surpassing existing lightweight and many heavy models in both accuracy and resource efficiency. Architectural insights from our model pave the way for real-time VSR on streaming data.
\section{Introduction}
\label{sec:intro}

Video Super-Resolution (VSR) aims to reconstruct high-resolution videos from their low-resolution counterparts. This task is crucial in a wide range of applications, including video streaming, gaming, mobile content delivery, and surveillance. By leveraging both spatial and temporal information, VSR enhances the visual quality of videos while maintaining temporal consistency. However, achieving high-quality super-resolution under computational constraints remains a persistent challenge, particularly in scenarios where resource efficiency is paramount.

%{model_performance_comparison_gflops_legend.png}
   
\begin{figure}[h]
    \centering
    \includegraphics[clip, trim=2cm 17.5cm 6cm 1.5cm, width=1\linewidth]{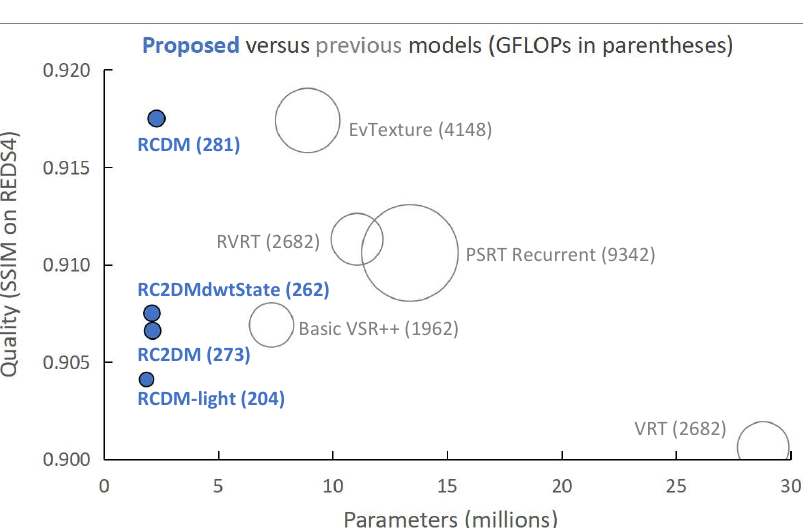}
    \caption{Comparison of video super-resolution (VSR) models: Proposed models (blue filled circles) reconstruct frames with competitive quality (SSIM on REDS4) when compared to previous models (grey hollow circles) with fewer trainable parameters and inference floating point operations per frames (GFLOPs in parenthesis for model labels and also coded as bubble size).}
    \label{fig:model_comparison}
\end{figure}

In recent years, deep learning-based approaches have shown remarkable progress in addressing the challenges of VSR. Convolutional architectures, such as video restoration framework with enhanced deformable networks (EDVR)~\cite{wang2019edvr} and BasicVSR++~\cite{chan2022basicvsr++}, have achieved notable success by employing deformable convolutions and residual networks to utilize motion information and recover fine details, respectively. Transformer-based models, including video restoration transformer (VRT) and recurrent video restoration transformer (RVRT), have further pushed the boundaries of VSR by capturing global temporal dependencies and improving long-term consistency\cite{liang2022vrt}\cite{liang2022rvrt}. However, their substantial parameter count and computational requirements limit their utility for deployment in setups with strict power, compute, and cost constraints.

VSR models have explored motion-aware architectures such as VESPCN\cite{vespcn} and EDVR\cite{wang2019edvr}, as well as methods tailored for compressed video super-resolution\cite{Guan2019MFQE2A}. These approaches aim to balance performance and efficiency, yet they often struggle with temporal coherence, handling large motion, and adapting to diverse real-world degradations—particularly in highly compressed or low-quality videos with unknown distortions.Furthermore, most existing methods focus solely on spatial and temporal domains, neglecting the potential benefits of incorporating frequency-domain representations for multi-scale feature extraction. This gap highlights the need for an innovative design that integrates these domains while maintaining computational efficiency.

In this work, we propose a novel lightweight VSR framework that combines convolutional efficiency with advanced spatiotemporal modeling to achieve state-of-the-art results with limited resources. Our architecture builds upon the strengths of convolutional networks, which excel at localized feature extraction, and enhances them with components tailored for challenges specific to VSR. To improve spatial and structural representation, we introduce a dual-pipeline design that incorporates 2D discrete wavelet transform (DWT) for multi-resolution and spatially sparse edge feature extraction and 3D convolutional residual blocks for spatio-temporal feature extraction. Additionally, we integrate deformable convolutions to ensure effective motion alignment across frames, which is crucial for maintaining temporal consistency. Furthermore, we leverage a memory tensor to enhance utilization and reconstruction of temporal coherence. Additionally, ConvNeXt blocks are employed to ensure efficient spatial feature extraction while maintaining a low parameter count and GFLOPs. 

This work addresses the critical challenge of balancing computational efficiency and restoration quality in VSR. By combining convolutional networks, wavelet-based feature extraction, and temporal memory mechanisms, our framework advances the state-of-the-art in lightweight VSR and paves the way for practical applications in resource-constrained scenarios.

Following are our key contributions:

\begin{itemize}
\item State-of-the-art reconstruction quality on REDS4\cite{Nah_2019_CVPR_Workshops_SR}: Our method achieves an SSIM of 0.9175, surpassing much heavier models, such as BasicVSR++\cite{chan2022basicvsr++} (0.9069) and RVRT (0.9113)\cite{liang2022rvrt}.

\item Low computational overhead compared to previous models: Transformer-based models such as VRT (30.7M parameters)\cite{liang2022vrt} and RVRT (11.06M parameters)\cite{liang2022rvrt} exhibit high computational costs, while our model delivers comparable or better performance at significantly lower resource requirements. The floating point operations required at test time are also from seven to 35 times lower than comparable models (see Figure~\ref{fig:model_comparison}).
    
\item Memory tensor enhancements: Our introduction of memory tensors plays a key role in preserving temporal consistency and significantly improving SSIM scores across multiple datasets. Compared to previous memory-based approaches, such as those using convolutional LSTM, our approach is significantly more compute efficient. %Our proposed memory tensor module efficiently retains past frame information while avoiding the memory overhead of recurrent-based methods. 
This approach enhances temporal stability without excessive model complexity.

\item Residual deformable convolutions: Our work builds on the strengths of deformable convolutions for motion alignment but incorporates residual deformable convolutions, which improve feature reuse and reduce the need for excessive parameterization.

\item Wavelet decomposition in 2D: We utilize the multi-resolution and energy compaction (sparsification) properties of 2D wavelets when applied to natural images to increase the spatial information utilization in our architecture.

\item Orchestration of multiple architectural features: While ideas such as 3D convolutions and residual blocks have been used in VSR before, our orchestration of multiple parallel branches that specialize in exploiting various aspects of spatio-temporal information available in videos is unique, which reflects in its results.

\end{itemize}
\section{Related Work}
\label{sec:related_work}

\textbf{Transformer-Based Video Super-Resolution (VSR) Models:} 
Transformer-based architectures, such as Video Restoration Transformer (VRT) cite{liang2022vrt} and Recurrent Video Restoration Transformer (RVRT)\cite{liang2022rvrt}, leverage global attention to model long-term temporal dependencies, achieving state-of-the-art performance in VSR. However, their high computational cost and memory footprint make them impractical for real-time applications or deployment on resource-constrained devices \cite{liu2025vsrdiff}. Progressive Sparse Representation Transformer (PSRT) \cite{psrt2023} improves efficiency using a recurrent design but still suffers from high latency. While these transformer-based methods achieve strong reconstruction quality, their computational and memory requirements limit their utility for deployment in low-resource settings.

\textbf{Lightweight VSR Models:} 
Several recent works have focused on reducing model complexity while maintaining competitive performance. Efficient Lightweight Network for Video Super-Resolution (ELNVSR) \cite{efficient2024lightweight} introduces a bidirectional alignment module and a multi-scale pyramid block to reduce redundancy and improve feature aggregation. Similarly, Information Refinement Network (IRN)\cite{zhang2023lightweight} employs ConvNeXt blocks with residual structures, effectively balancing parameter efficiency and accuracy. However, these models still struggle with motion compensation and temporal coherence, particularly in challenging real-world scenarios.  

A notable contribution in lightweight VSR is WaveMixSR\cite{wavemixsr2024}, which employs wavelet transforms to extract multi-scale representations. By leveraging joint spatial and frequency-domain processing, WaveMixSR enhances texture restoration while reducing redundant spatial computations. However, despite its advantages, it does not integrate effective memory mechanisms for long-term temporal dependencies, leading to frame inconsistencies in videos.

\textbf{Deformable Convolutions for Motion Alignment:}  
Deformable convolutions have proven highly effective in handling motion misalignment in VSR. Enhanced Deformable Video Restoration (EDVR) \cite{wang2019edvr} introduced deformable alignment layers that dynamically predict spatial offsets for better motion compensation. Basic Video Super-Resolution++ (BasicVSR++) \cite{chan2022basicvsrpp} further refined this approach by incorporating bidirectional propagation with deformable convolutions, leading to improved performance in dynamic scenes. However, while these methods enhance motion alignment, they often introduce additional parameters, increasing computational complexity.  

%Our work builds on the strengths of deformable convolutions but incorporates residual deformable convolutions, which improve feature reuse and reduce the need for excessive parameterization. [MOVED TO INTRO]

\textbf{Memory-Augmented Architectures for Temporal Consistency:}  
To address temporal inconsistencies in VSR, several models integrate memory-based architectures. RealBasicVSR \cite{chan2022investigating} introduces a dynamic refinement module that progressively reduces artifacts by leveraging long-term dependencies. Similarly, RVRT \cite{liang2022rvrt} incorporates memory mechanisms to improve video consistency across frames. However, these methods still rely on high computational budgets, making them less viable for low-resource applications.  

%Our proposed memory tensor module efficiently retains past frame information while avoiding the memory overhead of recurrent-based methods. This approach enhances temporal stability without excessive model complexity. [Moved to intro.]

\textbf{Efficient Upsampling and Multi-Scale Feature Extraction:}  
Efficient upsampling plays a crucial role in lightweight VSR models. Dynamic Context-Guided Upsampling (DCGU) \cite{huang2024dynamic} improves detail reconstruction by leveraging non-local sampling, enhancing fine-texture restoration with reduced computational cost. Meanwhile,wavelet-based methods have gained attention due to their ability to capture structural information across different frequency scales.  

We integrate \textbf{wavelet-based multi-scale feature extraction} in our framework to improve detail preservation while reducing redundant computations. Unlike traditional spatial convolutions, wavelet transformations enable a more efficient and parameter-effective method of capturing high-frequency details.

Building upon these advancements, we propose a convolution-based VSR model that achieves state-of-the-art performance while remaining computationally efficient. Unlike transformer-heavy methods, our approach incorporates:
\begin{itemize}
    \item \textbf{Residual Deformable Convolutions} for precise frame alignment with minimal overhead.
    \item \textbf{Memory Tensors} to enhance long-term temporal consistency without the need for excessive memory buffers.
    \item \textbf{Wavelet-Based Feature Extraction} for multi-scale detail enhancement while reducing computational redundancy.
    \item \textbf{Parameter-Efficient Design}, achieving superior performance on REDS4 and other benchmark datasets with only \textbf{2.3M parameters} and significantly fewer FLOPs compared to transformer-based approaches.
\end{itemize}

Our model demonstrates that by strategically integrating \textbf{memory mechanisms, residual networks, and deformable convolutions}, it is possible to \textbf{achieve high-quality VSR while maintaining low computational costs}, making it suitable for real-world applications such as video streaming, surveillance, and mobile devices.

\section{Proposed Method}
\label{sec:method}

We introduce \textbf{Residual ConvNeXt Deformable Convolution with Memory (RCDM)} -- a VSR architcture that is designed for computational efficiency while maintaining state-of-the-art performance. Unlike transformer-based methods, RCDM employs deformable convolutions for motion compensation, wavelet-based feature extraction for texture enhancement, and memory tensors for long-term temporal consistency.

\subsection{Problem formulation} Video super-resolution involves restoring high-resolution details from a set of degraded low-resolution frames. The degradation process can be modeled as:
\begin{equation}
    I_t = D(B(H_t)) + \eta_t,
\end{equation}
where \( B(\cdot) \) represents blur, \( D(\cdot) \) denotes downsampling, and \( \eta_t \) models additive noise. The primary goal of our network is to approximate the inverse mapping.

Given a sequence of $2N+1$ LR frames $\{I_{t-N}, ..., I_t, ..., I_{t+N}\}$, the goal of RCDM is to reconstruct the corresponding central HR frame $H_t$. The LR frame is loaded to have the size of $(2N+1,b,c,h,w)$, which corresponds to frames, batch, channels, height, width while the generated HR has a size $(1,b,c,4h,4w)$ corresponding to the batch of central frames. Mathematically, this can be formulated as:

\begin{equation}
    H_t, M_t = f_{\alpha}(I_{t-N}, ..., I_t, ..., I_{t+N}, M_{t-1}),
    \label{eq:vsr_function}
\end{equation}
where $f_{\alpha}$ is the learned VSR model parameterized by $\alpha$, and $M_{t-1}$ is a memory tensor storing information from previous frames to ensure temporal consistency. The model outputs both the estimated HR frame $H_t$ and the memory tensor to be utilized in the next frame $M_t$.

\subsection{Alignment Block}
A key challenge in VSR is motion alignment across consecutive frames to account for movements of the camera or objects. Instead of relying on explicit optical flow estimation, which can introduce artifacts, we utilize \textbf{Deformable Convolution for motion compensation} to dynamically adapt the receptive field and align adjacent frames with the reference frame. A standard convolution samples feature values from a fixed grid, but a deformable convolution introduces learned offsets to adjust sampling locations dynamically\cite{dai2017deformable}. Extending this to 3D allows the network to model both spatial and temporal motion, ensuring frame alignment without explicit motion estimation. The alignment block followed by feature extraction increases in channel depth while keeping the length and width as same as the LR frame (h,w)

\subsection{Residual Spatio-Temporal Feature Extraction}
Once aligned, the neighboring frame features are processed using a Residual 3D Convolution Block, which aggregates multi-frame information while preserving the structure of the reference frame. Residual connections extract spatio-temporal features, and ensures the preservation of base frame details.

\subsection{Wavelet-Based Multi-Scale Feature Extraction}
To enhance fine texture details, RCDM employs a \textbf{2D Discrete Wavelet Transform (DWT)} to decompose the central frame into frequency sub-bands capturing horizontal, vertical, and diagonal details. These sub-bands are processed independently and later fused to reconstruct high-quality textures. Due to the DWT, the size of features changes from $(h,w,c)$ to $(h/2,w/2,4c)$ without any loss of information. To enable feature fusion by concatenation, an up-sampling block prior to fusion is processed.

\subsection{Memory Mechanism for Temporal Consistency}
To maintain long-term consistency across video frames, we introduce a \textbf{memory tensor} $M_t$, which accumulates information over time using a residual update mechanism:

\begin{equation}
    M_t = \beta M_{t-1} + H_t^{\text{feat}},
    \label{eq:memory_update}
\end{equation}
where $\beta$ is a learnable weight controlling the influence of past frames, and $H_t^{\text{feat}}$ represents the refined features of the current frame. This allows the model to propagate previously learned details forward, preventing flickering artifacts in video sequences\cite{Chiche_2022_CVPR} \cite{chu2020learningtemporalcoherenceselfsupervision}.

\subsection{Super-Resolution and Upsampling}
The final stage refines the frame representation using \textbf{ConvNeXt Blocks}, which consist of depthwise convolutions and residual connections.\cite{liu2022convnext} The ConvNeXt blocks enable feature refinement while keeping computational cost low.

Finally, upsampling is performed using a \textbf{pixel shuffle} layer\cite{pixelshufflesr}, which efficiently rearranges feature maps to reconstruct the HR frame. 
In this block, HR upsampled versions are generated using pixel shuffle and ConvNeXT are combined for residual learning. This is then followed by learning from memory block to generate the expected final HR frame.

\begin{figure*}[h]
    \centering
    \includegraphics[width=0.95\linewidth]{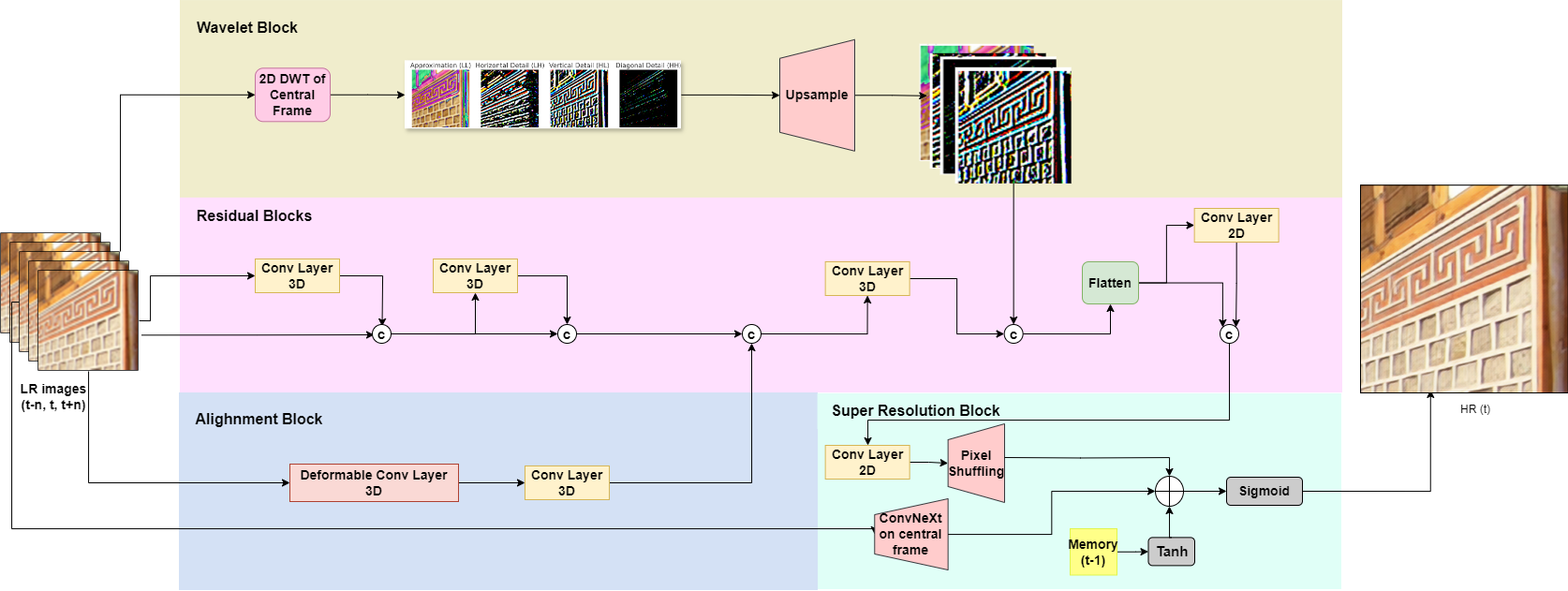}
    \caption{Overview of the proposed RCDM architecture. The model integrates deformable convolutions, wavelet decomposition, memory propagation, and ConvNeXt refinement for efficient video super-resolution.}
    \label{fig:rcdm_architecture}
\end{figure*}

The complete RCDM pipeline is shown in Figure~\ref{fig:rcdm_architecture}, which shows how it integrates:
\begin{enumerate}
    \item \textbf{Deformable Convolution for Motion Compensation}: Aligns neighboring frames dynamically.
    \item \textbf{Residual Feature Fusion}: Extracts spatio-temporal information while preserving details.
    \item \textbf{Wavelet-Based Processing}: Enhances multi-scale textures.
    \item \textbf{Memory Tensor}: Ensures temporal stability.
    \item \textbf{ConvNeXt Refinement and Upsampling}: Produces high-quality HR frames.
\end{enumerate}

At each timestep $t$, the model reconstructs $H_t$ while updating the memory for subsequent frames, ensuring smooth, artifact-free video super-resolution.

\subsection{Family of Architectures}
Our ResConvDefMem (RCDM) family of architectures consists of multiple variants, each offering a trade-off between accuracy and computational cost. 

\begin{enumerate}
    \item RCDM: This is our main model that uses deformable convolutions, wavelet transforms, and memory tensors for spatial and temporal consistency. 
    \item RCDM-light: This architecture consists of same components as our main architecture with the exception of using a lighter feature extraction layer.
    \item RC2DM: This variant uses early fusion to integrate contextual information from previous frames at an earlier stage. 
    \item RCDMdwtState: This variant uses wavelet-based memory integration for improved long-term temporal coherence. 
    \item RC2DMdwtState: This variant uses DWT-based feature enhancement of memory tensor for preserving fine-grained details.
\end{enumerate}
Table~\ref{tab:family} presents a detailed comparison.

\begin{table}[h]
\centering
\caption{Comparison of the RCDM family of architectures.}
\label{tab:family}
\resizebox{\columnwidth}{!}{%
\begin{tabular}{l r r r}
\hline
\textbf{Model} & \textbf{Params (M)} & \textbf{GFLOPs} & \textbf{SSIM} \\
\hline
RCDM-light       & 1.86  & 203.69  & 0.9041 \\
RC2DMdwtState    & 2.10  & 262.05  & 0.9075 \\
RC2DM            & 2.13  & 272.57  & 0.9066 \\
\textbf{RCDM}             & \textbf{2.30}  & \textbf{281.06}  & \textbf{0.9175} \\
RCDMdwtState     & 2.56  & 353.27  & 0.9129 \\
\hline
\end{tabular}%
}
\end{table}

\subsection{Performance and Computation Trade-offs}
The proposed RCDM model achieves high accuracy while remaining computationally efficient. Our approach leverages residual convolutions, deformable motion alignment, and memory-based feature propagation to maintain a balance between accuracy and efficiency.

We introduce a lightweight convolutional-based VSR framework that achieves high reconstruction accuracy with low computational complexity. Our model effectively integrates deformable convolutions for motion compensation, wavelet-based feature extraction for texture enhancement, and memory tensors for temporal coherence. Experimental results on the REDS BI dataset demonstrate that RCDM (ours) achieves a competitive SSIM of 0.9175 while maintaining only 2.3M parameters and 281 GFLOPs, making it well-suited for real-time video processing on edge devices.

\section{Experiments and Results}
\label{sec:experiments}

\subsection{Dataset and Training Setup}
We trained our model on the REalistic and Dynamic Scenes dataset (REDS) dataset \cite{nah2019ntire}, which is widely used for video super-resolution and restoration tasks. The dataset consists of 300 training sequences, each containing 100 frames with 720p resolution. To enhance the generalization capability of our model, we perform transfer learning on Vimeo90K \cite{xue2019video} and YouHQ datasets. These datasets offer diverse motion characteristics and scene complexities, making our model adaptable to real-world scenarios. For evaluation, we test our model on multiple benchmark datasets: REDS4\cite{nah2019ntire}, Vimeo90K-T \cite{xue2019video}, Vid4\cite{liu2013bayesian}, UDM10\cite{yi2019udm10}, and SPCMS\cite{tao2017spmc}. We compare structural similary index measure (SSIM) metric across the experiments as PSNR tends to prioritize global intensity over local detail matching.

\subsection{Implementation Details}
We use an input sequence of five frames, where the central frame is super-resolved. For real-time video super-resolution with a larger group of frames, our model can super-resolve the last frame of the sequence since it does not rely on future frames for reconstruction. Both approaches result in nearly the same quality.

During transfer learning, we retrain the memory tensor module and the ConvNeXt-based frame super-resolution block to adapt to the new datasets while keeping the feature extraction layers frozen. The model is implemented in PyTorch and trained on a single NVIDIA A100 GPU with 80 GB VRAM. Training is conducted using mixed-precision to optimize memory efficiency.

We use the AdamW optimizer with the following hyperparameters:  
Learning rate: \( 4 \times 10^{-4} \),  
\( \beta \): \( (0.9, 0.999) \),  
\( \epsilon \): \( 1 \times 10^{-8} \),  
Weight decay: \( 0.001 \).

For feature extraction, we leverage ConvNeXt blocks \cite{liu2022convnext} instead of transformers, ensuring reduced computational complexity while retaining long-range dependencies. Our model integrates deformable convolutions \cite{dai2017deformable} for motion compensation, wavelet-based multi-scale feature extraction \cite{huang2024dwtvsr}, and a memory tensor module for preserving temporal coherence across frames.

\subsection{Results}
Table~\ref{tab:vsr_comparison} presents a comparison of various video super-resolution methods across multiple datasets using the SSIM metric. Our proposed RCDM model achieves results comparable to the state-of-the-art on multiple datasets while using far fewer parameters (Table~\ref{tab:vsr_comparison}.

\begin{table*}[h]
\caption{Comparison of VSR models across multiple datasets (Other method's metrics are taken from their respective papers.)}
\centering
 \begin{adjustbox}{width=0.95\textwidth}
\begin{tabular}{l|c|c|c|c|c|c|c}
\hline
\textbf{Dataset} & \textbf{\#Params (M)} & \multicolumn{3}{c|}{\textbf{BI Degradation (Bicubic Downsampling)}} & \multicolumn{3}{c}{\textbf{BD Degradation (Blur-Downsampling)}} \\
\hline
 &  & \textbf{REDS4 (RGB)} & \textbf{Vimeo90K-T} & \textbf{Vid4} & \textbf{UDM10} & \textbf{Vimeo90K-T} & \textbf{Vid4} \\
\hline

Bicubic & - & 0.7292 & - & - & - & - & - \\
TOFlow \cite{xue2019tof} & - & 0.7990 & 0.9054 & - & - & - & - \\
EDVR \cite{wang2019edvr} & 20.6 & 0.8800 & 0.9489 & - & - & - & - \\

BasicVSR++ \cite{chan2022basicvsrpp} & 7.32 & 0.9069 & 0.9500 & 0.8400 & 0.9722 & 0.9550 & 0.8753 \\
EvTexture\cite{evtexture2023} & 8.90 & 0.9174 & 0.9544 & 0.8909 & - & - & - \\
RVRT \cite{liang2022rvrt} & 11.06 & 0.9113 & 0.9527 & 0.8462 & 0.9729 & 0.9576 & 0.8810 \\
PSRT Recurrent\cite{psrt2023} & 13.37 & 0.9106 & 0.9536 & 0.8485 & - & - & - \\
VRT \cite{liang2022vrt} & 28.80 & 0.8795 & 0.9530 & 0.8425 & 0.9737 & 0.9584 & 0.9006 \\
\hline
RCDM (Ours) & 2.30 & \textbf{0.9175} & 0.9212 & 0.7766 & 0.9502 & 0.9194 & 0.8541 \\

\hline
\end{tabular}
\end{adjustbox}

\label{tab:vsr_comparison}
\end{table*}

Our method demonstrates impressive performance on the REDS4 dataset, surpassing BasicVSR++ (0.9069) and RVRT (0.9113). Additionally, it remains competitive on Vimeo90K-T and UDM10, despite transformer-based models like VRT and RVRT showing strong results. Notably, our approach achieves this with fewer parameters and inference compute (Figure~\ref{fig:model_comparison}).

\begin{figure*}
    \centering
    \subfloat{
        % Row 1: Cactus Scene
        \begin{minipage}{0.2\textwidth}
            \centering
            \begin{overpic}[width=\textwidth]{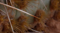}
                \put(5,5){\includegraphics[width=0.40\textwidth]{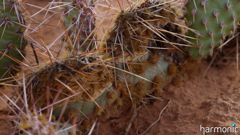}}
            \end{overpic}
        \end{minipage}
        \begin{minipage}{0.2\textwidth}
            \centering
            \includegraphics[width=\textwidth]{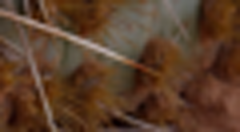} % Bicubic
        \end{minipage}
        \begin{minipage}{0.2\textwidth}
            \centering
            \includegraphics[width=\textwidth]{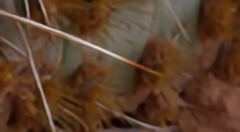} % Model Output
        \end{minipage}
    }\\

    \subfloat{
        % Row 2: Car Scene
        \begin{minipage}{0.2\textwidth}
            \centering
            \begin{overpic}[width=\textwidth]{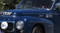}
                \put(5,5){\includegraphics[width=0.40\textwidth]{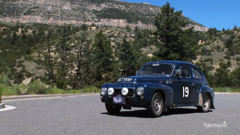}}
            \end{overpic}
        \end{minipage}
        \begin{minipage}{0.2\textwidth}
            \centering
            \includegraphics[width=\textwidth]{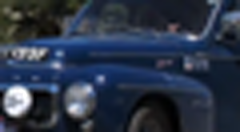} % Bicubic
        \end{minipage}
        \begin{minipage}{0.2\textwidth}
            \centering
            \includegraphics[width=\textwidth]{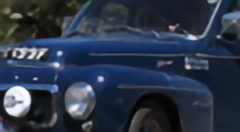} % Model Output
        \end{minipage}
    }\\

    \subfloat{
        % Row 5: Building Scene
        \begin{minipage}{0.2\textwidth}
            \centering
            \begin{overpic}[width=\textwidth]{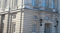}
                \put(5,5){\includegraphics[width=0.40\textwidth]{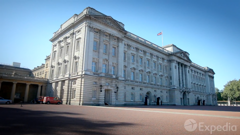}}
            \end{overpic}
        \end{minipage}
        \begin{minipage}{0.2\textwidth}
            \centering
            \includegraphics[width=\textwidth]{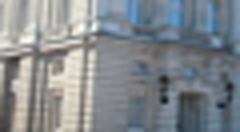} % Bicubic
        \end{minipage}
        \begin{minipage}{0.2\textwidth}
            \centering
            \includegraphics[width=\textwidth]{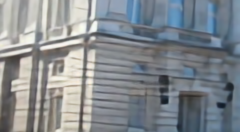} % Model Output
        \end{minipage}
    }

    \caption{Patch-wise comparison across different samples of the SPMCS dataset. First column: input LR frame (with overlaid patch), Second: bicubic upsampling, Third: our model's output.}
    \label{fig:patch_comparison}
\end{figure*}

To further demonstrate the effectiveness of our model, we present qualitative comparisons against bicubic upsampling across diverse datasets. Figure \ref{fig:patch_comparison} presents the patch results in different scenes adopted from the SPCMS\cite{tao2017spmc} dataset, where fit can be seen that fine details are recreated with sharpness and without artifacts.

\begin{figure*}
    \centering

    % ---- First Set ----
    \begin{minipage}{0.19\textwidth}
        \centering
        \textbf{Low Resolution}
        \begin{overpic}[width=\textwidth]{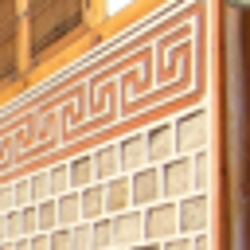} % LR Input
            \put(5,5){\includegraphics[width=0.60\textwidth]{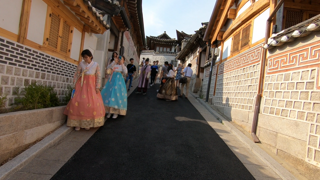}} % Enlarged Overlay
        \end{overpic}
    \end{minipage}
    \begin{minipage}{0.19\textwidth}
        \centering
        \textbf{Bicubic}
    \includegraphics[width=\textwidth]{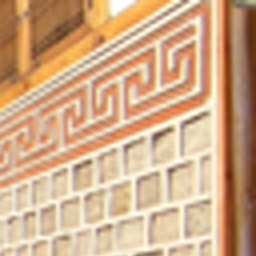} % Bicubic Upsampling
    \end{minipage}
    \begin{minipage}{0.19\textwidth}
        \centering
    \textbf{RVRT}    \includegraphics[width=\textwidth]{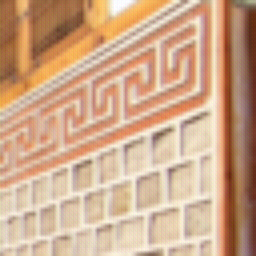} % RVRT Output

    \end{minipage}
    \begin{minipage}{0.19\textwidth}
        \centering
        \textbf{BasicVSR++}
        \includegraphics[width=\textwidth]{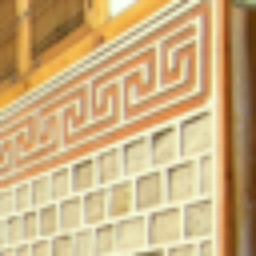} % BasicVSR++
    \end{minipage}
    \begin{minipage}{0.19\textwidth}
        \centering
        \textbf{Ours}
        \includegraphics[width=\textwidth]{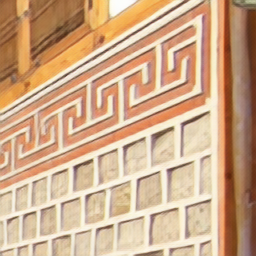} % Our Model Output
    \end{minipage}

    \vspace{10pt} % Add spacing between sets

    % ---- Second Set ----
    \begin{minipage}{0.19\textwidth}
        \centering
        \begin{overpic}[width=\textwidth]{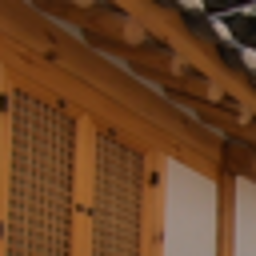} % LR Input
            \put(5,5){\includegraphics[width=0.60\textwidth]{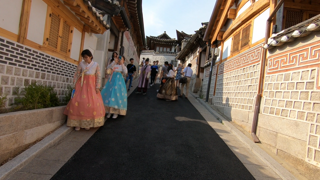}} % Enlarged Overlay
        \end{overpic}
    \end{minipage}
    \begin{minipage}{0.19\textwidth}
        \centering
        \includegraphics[width=\textwidth]{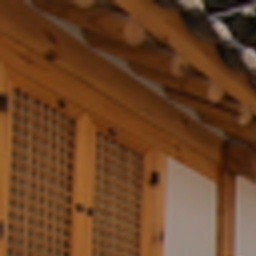} % Bicubic Upsampling
    \end{minipage}
    \begin{minipage}{0.19\textwidth}
        \centering
        \includegraphics[width=\textwidth]{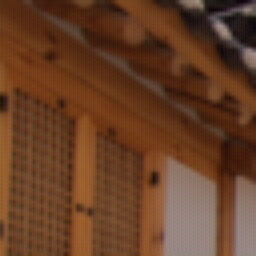} % RVRT Output

    \end{minipage}
    \begin{minipage}{0.19\textwidth}
        \centering
        \includegraphics[width=\textwidth]{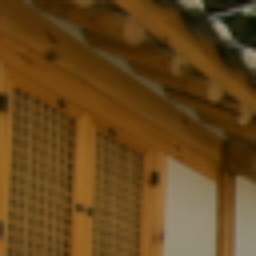} % BasicVSR++
    \end{minipage}
    \begin{minipage}{0.19\textwidth}
        \centering
        \includegraphics[width=\textwidth]{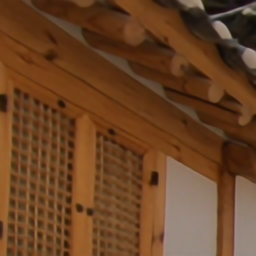} % Our Model Output
    \end{minipage}

    \vspace{10pt} % Add spacing between sets
    % ---- Fourth Set ----
    \begin{minipage}{0.19\textwidth}
        \centering
        \begin{overpic}[width=\textwidth]{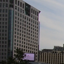} % LR Input
            \put(5,5){\includegraphics[width=0.60\textwidth]{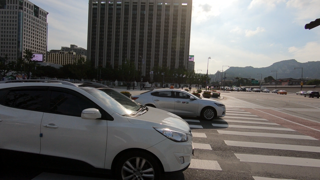}} % Enlarged Overlay
        \end{overpic}
    \end{minipage}
    \begin{minipage}{0.19\textwidth}
        \centering
        \includegraphics[width=\textwidth]{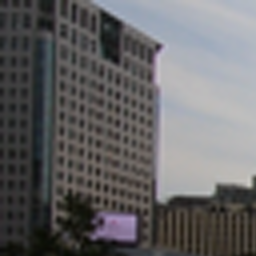} % Bicubic Upsampling
    \end{minipage}
    \begin{minipage}{0.19\textwidth}
        \centering
        \includegraphics[width=\textwidth]{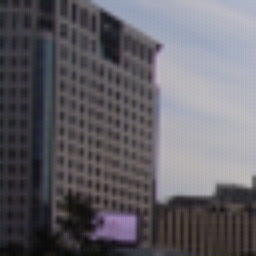} % RVRT Output
    \end{minipage}
    \begin{minipage}{0.19\textwidth}
        \centering
        \includegraphics[width=\textwidth]{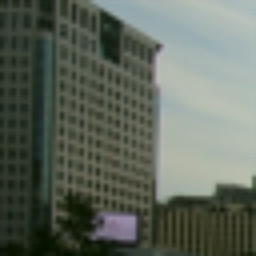} % BasicVSR++
    \end{minipage}
    \begin{minipage}{0.19\textwidth}
        \centering
        \includegraphics[width=\textwidth]{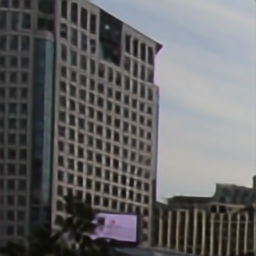} % Our Model Output
    \end{minipage}

    \caption{Patch-wise comparison across different samples. Each row represents a different case:
    (1) Low-Resolution input with an enlarged overlaid patch, 
    (2) Bicubic upsampling, 
    (3) RVRT output, 
    (4) BasicVSR++ output, 
    (5) Our output.}
    \label{fig:patch_comparison_all}
\end{figure*}

\subsection{Analysis of Results}
Our model demonstrates superior detail preservation compared to Bicubic Upsampling, RVRT, and BasicVSR++. The fine structures in architectural patterns, edges, and textures are sharply reconstructed, while other methods exhibit blurring or texture loss. The inclusion of memory tensors and wavelet-based feature extraction ensures that the generated frames maintain both temporal consistency and high-frequency details, outperforming transformer-based approaches that often introduce artifacts or over-smoothing. These results confirm the effectiveness of our method in real-world scenarios where fine-grained structures and text clarity are critical.

Our model performs well on the the REDS4 dataset as it is a long sequence along with dynamic motion. Table \ref{tab:motion_comparison} shows the nature of the datasets under consideration.
Our model performs with dataset with high motion similar to REDS

\begin{table*}[h]
    \centering
    \caption{Comparison of Motion Complexity Across Datasets}
    \renewcommand{\arraystretch}{0.8} % Increase row height for readability
    \setlength{\tabcolsep}{10pt} % Adjust column spacing
    \footnotesize % Reduce font size for compact table
    \begin{tabular}{lccc}
        \toprule
        \textbf{Dataset} & \textbf{Motion Complexity} & \textbf{Frame Rate} & \textbf{Key Challenges} \\
        \midrule
        \multirow{2}{*}{\textbf{REDS4\cite{nah2019ntire}}}  
        & \multirow{2}{*}{\textbf{Very High}} & \multirow{2}{*}{24 fps} 
        & - Fast-moving objects, occlusions. \\
        &  &  & - Large scene variations. \\
        \midrule
        
        \multirow{2}{*}{\textbf{UDM10\cite{yi2019udm10}}}  
        & \multirow{2}{*}{\textbf{Moderate - High}} & \multirow{2}{*}{30 fps} 
        & - Real-world compression artifacts. \\
        &  &  & - Camera jitter, lossy motion. \\
        \midrule
        
        \multirow{2}{*}{\textbf{Vimeo-90K\cite{xue2019video}}}  
        & \multirow{2}{*}{\textbf{Moderate}} & \multirow{2}{*}{~30 fps} 
        & - Well-curated natural motion. \\
        &  &  & - Fewer occlusions, predictable motion. \\
        \midrule
        
        \multirow{2}{*}{\textbf{Vid4\cite{liu2013bayesian}}}  
        & \multirow{2}{*}{\textbf{Low}} & \multirow{2}{*}{24 fps} 
        & - Mostly slow camera panning. \\
        &  &  & - Minimal motion complexity. \\
        \bottomrule
    \end{tabular}
    \label{tab:motion_comparison}
\end{table*}

The qualitative comparisons in Figure \ref{fig:patch_comparison_all} highlight the superiority of our model in capturing fine details while maintaining computational efficiency. Below are the key observations:
\textbf{Preservation of Fine Details}:  
- Our model effectively restores sharp textures and edges, as observed in cactus spikes, car headlights, and building edges.
- Unlike bicubic interpolation, which produces blurry and pixelated results, our method reconstructs clear and natural textures.
\textbf{High-Quality Super-Resolution Without Transformers}:  
- Unlike transformer-based approaches, which introduce global self-attention, our model preserves local structures using ConvNeXt blocks and deformable convolutions.
- Transformers are beneficial for high-level vision tasks (e.g., segmentation) but are unnecessary for low-level vision tasks like super-resolution, where local feature extraction is more effective.
\textbf{Lower Computational Complexity Without Sacrificing Quality}:  
- By leveraging memory mechanisms and efficient convolutions, our approach achieves comparable or superior quality with significantly lower computational overhead than transformer-based methods.
- This makes our model highly practical for real-time applications, including edge devices and video processing.
\textbf{Robust Performance Across Textures and Patterns}:  
- Our model generalizes well to natural textures, urban environments, and structured patterns, delivering consistent performance across diverse datasets.

Our results confirm that global self-attention is not always necessary for video super-resolution, and carefully crafted local architectures can achieve high-quality reconstruction with lower computational costs. Our model is a highly practical alternative to transformer-based VSR approaches particularly for edge devices.

\section{Ablation Study}
\label{sec:ablation}
To systematically evaluate the contributions of different components in our proposed architecture, we conducted an extensive ablation study. This study assesses the impact of architectural modifications on model complexity (measured in parameters) and reconstruction accuracy (measured in SSIM). 

\begin{figure*}
    \centering
    \subfloat[SSIM Improvement using Memory Mechanism]{%
        \includegraphics[width=0.48\textwidth]{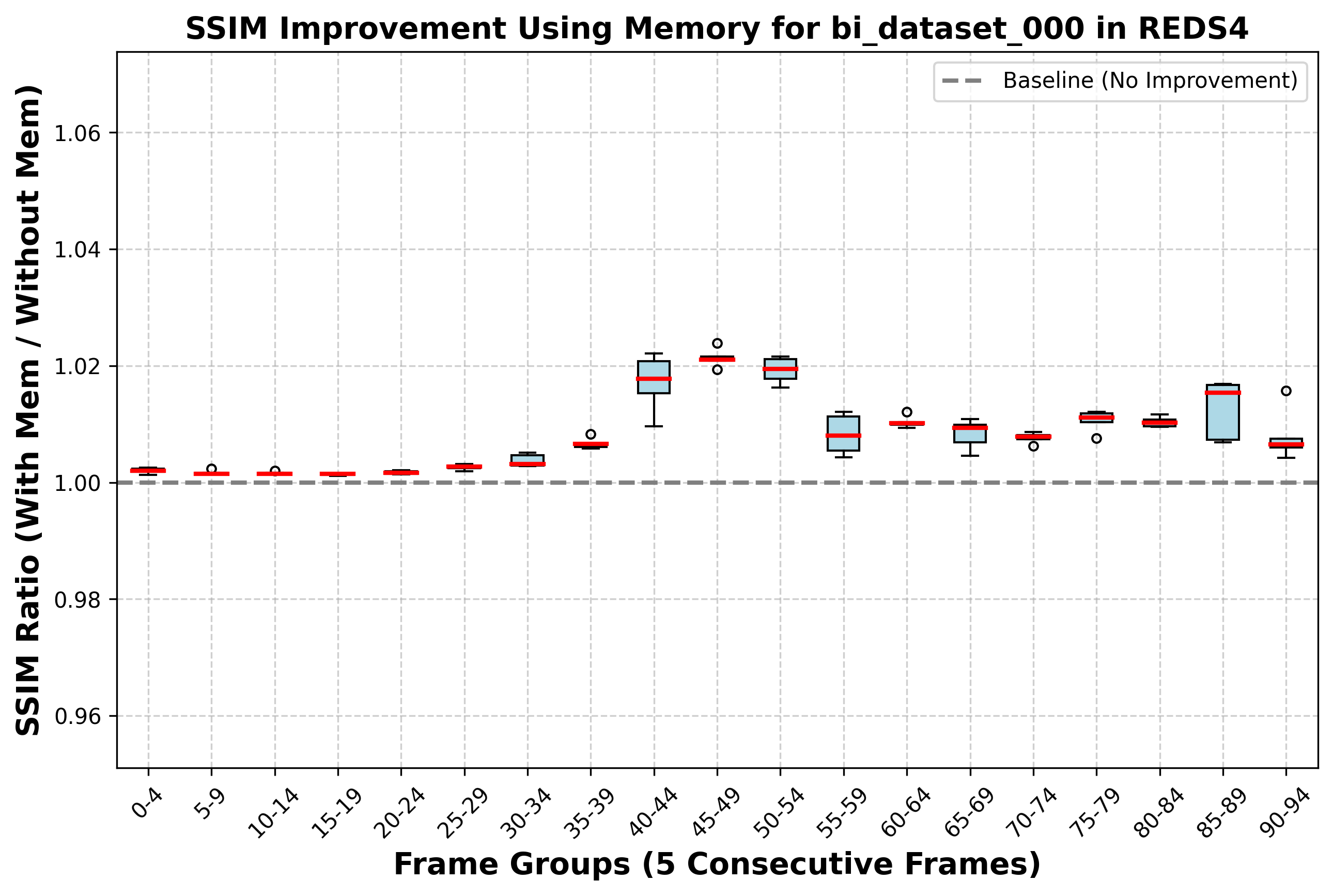}
        \label{fig:ssim_memory}
    }
    \hfill
    \subfloat[SSIM Improvement using Wavelet-Based Feature Extraction]{%
        \includegraphics[width=0.48\textwidth]{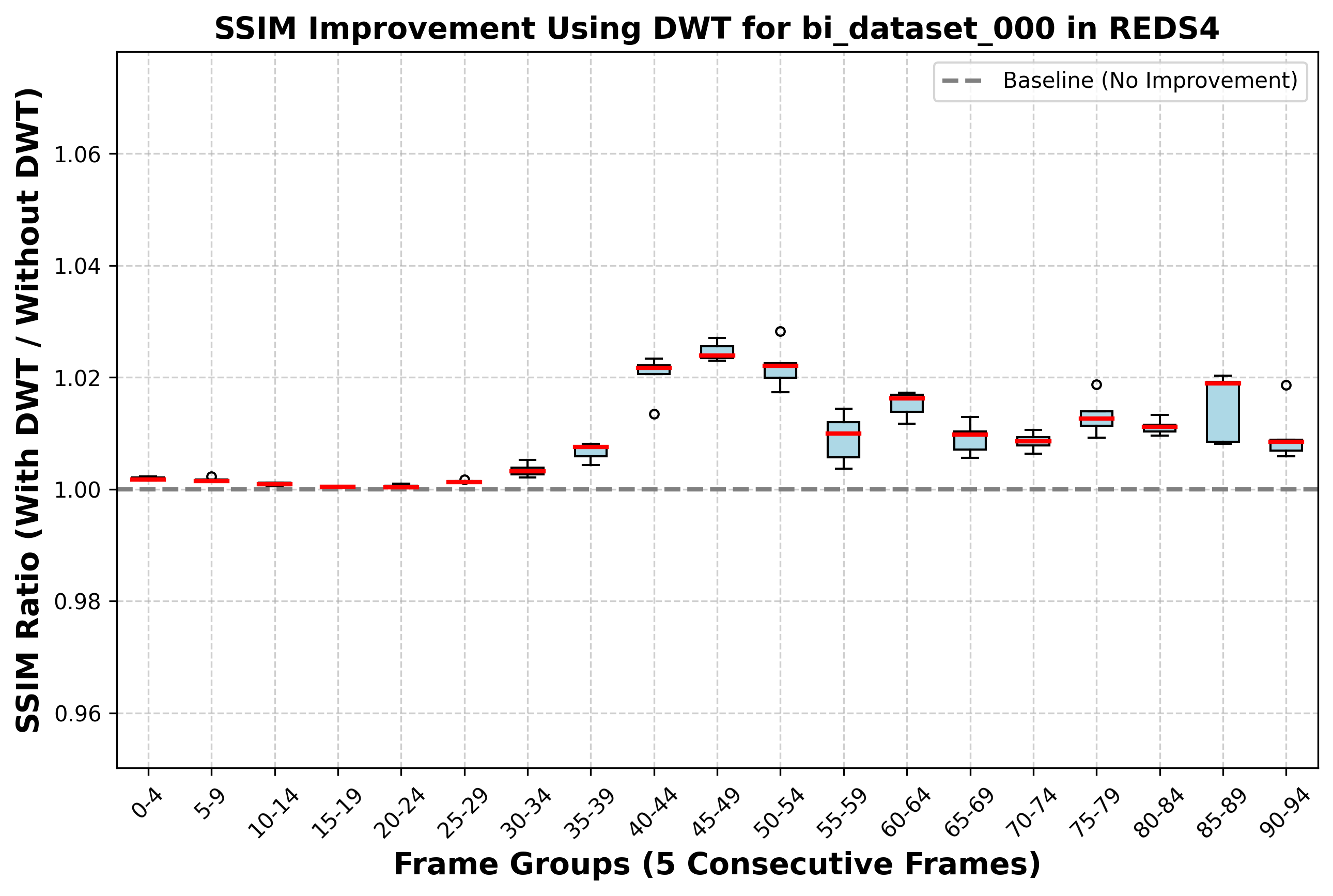}
        \label{fig:ssim_wavelet}
    }
    \caption{Ablation Study: (a) Impact of memory mechanism and (b) Impact of wavelet-based feature extraction on SSIM performance.}
    \label{fig:ssim_ablation}
\end{figure*}

\textbf{Impact of Memory and Wavelet Transform:}
The ablation study evaluates the effectiveness of two key components in our Residual ConvNeXt Deformable Memory (RCDM) model: memory mechanism and wavelet-based feature extraction. Figure~\ref{fig:ssim_ablation} presents the SSIM ratio for both components (using an architecture without these components as baseline), showing their contribution to enhancing video super-resolution quality on a subset of REDS4 dataset.

Figure~\ref{fig:ssim_memory} shows that memory propagation significantly improves temporal consistency, especially in later frames as more evidence to resolve patches is accumulated by the memory. Thus, the memory mechanism helps retain long-range dependencies, reducing flickering and enhancing detail retention in dynamic scenes. 

Figure~\ref{fig:ssim_wavelet} shows that the wavelet decomposition improves high-frequency detail preservation, leading to better texture reconstruction. Unlike standard convolutions, wavelet transforms effectively separate structural information across different frequency bands, enabling finer detail recovery. This results in higher SSIM ratios, especially in complex background regions.
Frame-to-frame consistency benefits from memory propagation, particularly in high-motion sequences.
Wavelets enhance texture reconstruction: By leveraging frequency decomposition, wavelet-based models preserve edges and fine details more effectively.

In fact, the memory and wavelet components complement each other. Memory helps smooth temporal transitions, while wavelets refine structural integrity. Their combined use yields optimal results in video super-resolution.

We also performed experiments in YCbCr space instead of RGB space to compare model performance. However, for our dataset of interest (REDS4), the model gave better performance in the RGB space.

\section{Conclusion and Future Work}
\label{sec:conclusion}

In this work, we introduced ResConvDefMem (RCDM), a novel lightweight and computationally efficient video super-resolution (VSR) model designed to bridge the gap between model complexity and reconstruction quality. By leveraging a combination of deformable convolutions, wavelet-based multi-scale feature extraction, and a memory tensor mechanism, our approach enhances both spatial and temporal coherence while maintaining a significantly lower computational cost compared to transformer-based methods.

Our experiments demonstrate that RCDM achieves reconstruction quality comparable to the state-of-the-art on several datasets and surpassing SOTA on the REDS4 dataset with an SSIM of 0.9175 when comapred to both lightweight VSR models and high-resource transformer-based architectures such as VRT and RVRT. 
Furthermore, our qualitative and quantitative results validate the  generalizability across diverse datasets. By prioritizing convolutional efficiency over transformer-based global attention, our model offers a practical solution for real-time applications such as video streaming, surveillance, and edge computing.

 While memory-based feature propagation improves temporal consistency, further enhancements are needed to ensure finer detail preservation and structure reconstruction in challenging scenarios. We can refine RCDM, perhaps using locally limited self-attention (e.g. that in SWIN transformer~\cite{liu2021swin}, to maximize both perceptual quality and computational efficiency, making it an optimal solution for next-generation video enhancement applications.

\bibliographystyle{plain}
\bibliography{main} % Use your .bib file


\end{document}